\documentclass[letterpaper, 10 pt, conference]{ieeeconf}  % Comment this line out if you need a4paper

\IEEEoverridecommandlockouts                              % This command is only needed if 
\usepackage{amsmath,amsfonts}
\usepackage[ruled,vlined,linesnumbered]{algorithm2e}
\usepackage{array}
\usepackage{subcaption}
\usepackage{textcomp}
\usepackage{dirtytalk}
\usepackage{stfloats}
\usepackage{url}
\usepackage{verbatim}
\usepackage{graphicx}
\usepackage{cite}
\usepackage{wrapfig}
\usepackage{url}            % simple URL typesetting
\usepackage{booktabs}       % professional-quality tablesmath symbols
\usepackage{amssymb}
\usepackage{nicefrac}       % compact symbols for 1/2, etc.
\usepackage{microtype}      % microtypography
\usepackage{tikz}
\usepackage{soul}
\usepackage{mathtools}
\newtheorem{remark}{Remark}

\newcommand{\bG}{{\mathbf G}}

\newcommand{\bX}{{\mathbf X}}

\newcommand{\bL}{{\mathbf L}}

\newcommand{\pluseq}{\mathrel{+}=}

\usepackage{etoolbox}
\makeatletter
\patchcmd{\@begintheorem}{\textit}{\textbf}{}{}
\makeatother

\begin{document}
\title{Accelerating Sampling-Based Control via  Learned Linear Koopman Dynamics}

\author{Wenjian Hao, Yuxuan Fang, Zehui Lu, and Shaoshuai Mou
\thanks{This material is based upon work supported by the Defense Advanced Research Projects Agency (DARPA) under of the Learning Introspective Control (LINC) project (grant no. N65236-23-C-8012) Any opinions, ﬁndings and conclusions or recommendations expressed in this material are those of the author(s) and do not necessarily reﬂect the views of the DARPA or the U.S. Government.}
\thanks{W. Hao, Y. Fang, and S. Mou are with the School of Aeronautics and Astronautics, Purdue University, West Lafayette, IN 47907, USA. { (Email: \texttt{\{hao93, fang394, mous\}@purdue.edu}) }}
\thanks{Z. Lu is an independent researcher. (Email:  \texttt{zehuilu789@gmail.com})}
}

% \author{
% \thanks{The video is available in supplementary files.}
% }




\maketitle

\begin{abstract}
This paper presents an efficient model predictive path integral (MPPI) control framework for systems with complex nonlinear dynamics. To improve the computational efficiency of classic MPPI while preserving control performance, we replace the nonlinear dynamics used for trajectory propagation with a learned linear deep Koopman operator (DKO) model, enabling faster rollout and more efficient trajectory sampling. The DKO dynamics are learned directly from interaction data, eliminating the need for analytical system models. The resulting controller, termed MPPI-DK, is evaluated in simulation on inverted pendulum swing-up and stabilization and surface vehicle navigation tasks, and validated on hardware through reference-tracking experiments on a quadruped robot. Experimental results demonstrate that MPPI-DK achieves control performance close to MPPI using true dynamics while substantially reducing computational cost, enabling efficient real-time control on robotic platforms. These results indicate that MPPI-DK provides an effective surrogate for sampling-based control when nonlinear dynamics propagation limits the achievable control frequency.
\end{abstract}

\section{Introduction}

The control of robotic systems with nonlinear, high-dimensional dynamics remains challenging, particularly for rapid maneuvers and real-time operation. Although model predictive control (MPC) can explicitly handle state and input constraints while optimizing performance over a receding horizon \cite{grune2016nonlinear, rawlings2020model}, its use at high control frequencies is often limited by repeated online optimization, nonlinear dynamics propagation, and occasional infeasibility. These computational demands can limit real-time robotic deployment. To improve scalability for highly nonlinear systems, model predictive path integral (MPPI) control approximates optimal control updates through Monte Carlo trajectory sampling rather than deterministic online optimization \cite{williams2016aggressive}. This sampling-based formulation accommodates nonlinear dynamics and nonconvex costs, supports parallel computation, and has been successfully applied to autonomous driving \cite{williams2016aggressive}, aerial robotics \cite{minavrik2024model}, and legged locomotion \cite{alvarez2025real}.

A key limitation of MPPI is the repeated propagation of nonlinear dynamics during trajectory sampling, which can restrict control frequency and scalability for computationally demanding systems. Data-driven surrogate models, particularly deep neural networks (DNNs)  \cite{williams2017information}, can approximate complex dynamics but may still incur substantial cost when evaluated repeatedly within sampling-based control. An alternative perspective of data-driven dynamics learning is using the Koopman operator theory, which represents nonlinear dynamics as linear evolution in a lifted state space \cite{mezic2015applications,mauroy2016linear}. Extended dynamic mode decomposition (EDMD) implements this idea using manually selected lifting functions \cite{korda2018convergence}, whereas deep Koopman operator (DKO) methods learn these lifting functions directly from data using DNNs \cite{lusch2017data,dk,hao2024deep}. DKO models have been integrated with MPC and model-based reinforcement learning \cite{korda2018linear,hao2025optimalcontrolnonlinearsystems}. Their linear structure enables efficient state propagation through matrix multiplications, making them suitable for MPPI control.

Although DKO–based methods have been extensively studied for dynamics identification and prediction, their use to accelerate real-time sampling-based control remains relatively underexplored. Motivated by this gap, we integrate linear DKO dynamics with MPPI to improve the efficiency of sampling-based optimal control for complex nonlinear robotic systems. The central idea is to replace repeated nonlinear dynamics evaluations during MPPI rollouts with linear propagation in the learned Koopman lifted space, while retaining compatibility with stochastic sampling and general nonconvex cost functions. The main contributions are summarized as follows:
\begin{itemize}
\item \textit{DKO-accelerated MPPI:} We develop an MPPI controller based on learned linear DKO dynamics. By propagating lifted states through linear operator evaluations rather than repeatedly evaluating nonlinear DNN models during trajectory rollouts, the proposed formulation substantially reduces computational cost while retaining compatibility with stochastic sampling and nonconvex cost functions.

\item \textit{Simulation and hardware validation:} We evaluate MPPI-DK on an inverted pendulum, a surface vehicle, and a Unitree Go1 quadruped. The results demonstrate favorable control computation trade-offs relative to MPPI using the true dynamics and MPC using the same learned model, together with substantial GPU speedups.
\end{itemize}
The remainder of the paper is organized as follows. Section~\ref{problem_form} formulates the problem and introduces the necessary preliminaries. Section~\ref{MaRe} presents the proposed framework. Section~\ref{numerical_sim} provides numerical simulation results, and Section~\ref{exp} demonstrates the approach on a quadruped robot platform. Finally, Section~\ref{conclusion} concludes the paper.

\textbf{Notation.}  $\lVert \cdot \lVert$ denotes the Euclidean norm. $\mathbb{R}_+$ denotes a set of all positive real numbers. For a matrix $A\in\mathbb{R}^{n\times m}$, $A'$ denotes its transpose and $A^\dagger$ denotes its Moore–Penrose pseudoinverse.
 
\section{The Problem and Preliminaries}\label{problem_form}
This section first formulates the problem addressed in this paper. It then reviews the DKO framework for data-driven dynamics approximation and introduces the classic MPPI control method as a baseline solution to the formulated problem.
\subsection{Problem Formulation}
Consider the following discrete-time nonlinear system:
\begin{equation}\label{eq_dyn}
    \mathbf{x}(t+1) = \boldsymbol{f}(\mathbf{x}(t), \mathbf{v}(t)),
\end{equation}
where $t=0,1,2,\cdots$ denotes the time index, $\mathbf{x}(t)\in\mathcal{X}\subset\mathbb{R}^n$ is the system state, and $\boldsymbol{f}:\mathcal{X}\times\mathcal{V}\rightarrow\mathcal{X}$ is a time-invariant nonlinear mapping. The applied control input $\mathbf{v}(t)\in\mathcal{V}\subset\mathbb{R}^m$ is subject to disturbances and is given by \[\mathbf{v}(t) = \mathbf{u}(t) + \delta\mathbf{u}(t),\] where $\mathbf{u}(t)\in\mathcal{U}\subset\mathbb{R}^m$ is the nominal control input and $\delta\mathbf{u}(t)\sim\mathcal{N}(0, \Sigma)$ is zero-mean Gaussian noise with constant covariance $\Sigma\in\mathbb{R}^{m\times m}$. 

We assume that the nonlinear dynamics in \eqref{eq_dyn} can be represented by a finite-dimensional linear Koopman dynamics, given by \begin{equation}\label{eq_approx_dko}
\begin{aligned}
    \mathbf{x}(t+1) = C^*\Big(A^* \boldsymbol{g}(\mathbf{x}(t),\boldsymbol{\theta}^*) + B^* \mathbf{v}(t)\Big), 
\end{aligned}
\end{equation} where $A^*\in\mathbb{R}^{r\times r}, B^*\in\mathbb{R}^{r\times m}, C^*\in\mathbb{R}^{n\times r}$, and $\boldsymbol{\theta}^*\in\mathbb{R}^p$ are constant matrices and the parameter vector to be determined, respectively. The function $\boldsymbol{g}(\cdot,\boldsymbol{\theta}^*): \mathcal{X}\rightarrow\mathbb{R}^r$ with $r\geq n$ is assumed to be Lipschitz continuous and is typically parameterized by a DNN.

The \textbf{problem of interest} is twofold: (i) to estimate the DKO dynamics in \eqref{eq_approx_dko} from state–input data generated by the original system \eqref{eq_dyn}, and (ii) to utilize the learned model to compute a finite-horizon control sequence \[\mathbf{U} = [\mathbf{u}(t), \mathbf{u}(t+1), \cdots, \mathbf{u}(t+T-1)]\in\mathbb{R}^{m\times T}\] that minimizes an expected cost. This leads to the following finite-horizon stochastic optimal control problem: \begin{equation}\label{eq_dk_mppi_optim}
    \begin{aligned}
        \min_{\mathbf{U}} &J(\mathbf{\hat x}, \mathbf{u})\! =\! \mathbb{E}\Big[\sum_{s=t}^{t+T-1}\!\! \hat{c}(\mathbf{\hat x}(s), \mathbf{u}(s),\delta\mathbf{u}(s)) \!+ \! \phi(\mathbf{\hat x}(t+T)) \Big]\\
    \text{s.t.} \quad &\mathbf{\hat x}(s+1) =  C^*\Big(A^* \boldsymbol{g}(\mathbf{\hat x}(s),\boldsymbol{\theta}^*) + B^* \mathbf{v}(s)\Big),   \\
   &\mathbf{\hat x}(t) = \mathbf{x}_t,\ \mathbf{\hat x}(s)\in\mathcal{X},\  \mathbf{u}(s)\in\mathcal{U},\ \delta\mathbf{u}(s)\sim\mathcal{N}(0, \Sigma),
    \end{aligned}
\end{equation}
where $\hat{c}:\mathcal{X}\times\mathcal{U}\times\mathbb{R}^m\rightarrow\mathbb{R}$ and $\phi:\mathcal{X}\rightarrow\mathbb{R}$ denote the stage and terminal cost functions, respectively.

\subsection{Dynamics Learning using DKO}
One approach to approximate \eqref{eq_approx_dko} is through the DKO framework. In the noise-free case, i.e.,  $\delta\mathbf{u}(s)\equiv 0$, the system dynamics in \eqref{eq_approx_dko} can be rewritten in a lifted linear representation as
\begin{align}
    \boldsymbol{g}(\mathbf{x}(t+1),\boldsymbol{\theta}^*) &= A^* \boldsymbol{g}(\mathbf{x}(t),\boldsymbol{\theta}^*) + B^* \mathbf{u}(t), \label{eq_lift_koopman} \\
\mathbf{x}(t+1) &= C^* \boldsymbol{g}(\mathbf{x}(t+1),\boldsymbol{\theta}^*), \label{eq_x_koopman}
\end{align}
where \eqref{eq_lift_koopman} describes the evolution in the lifted space and \eqref{eq_x_koopman} assumes the existence of a linear mapping between $\mathbf{x}(t+1)$ and its lifted states $\boldsymbol{g}(\mathbf{x}(t+1),\boldsymbol{\theta}^*)$. A detailed estimation error analysis of \eqref{eq_approx_dko} can be found in \cite{hao2024deep}. To approximate \eqref{eq_lift_koopman}-\eqref{eq_x_koopman}, consider a dataset of observed state–input–next-state tuples
\begin{equation}
    \mathcal{D} = \cup_{i=1}^{M}\{(\mathbf{x}_i,\mathbf{u}_i, \mathbf{x}_i^+)\},\nonumber
\end{equation} where $\mathbf{x}_i^+$ denotes the successor state obtained by applying input $\mathbf{u}_i$ to the true dynamics \eqref{eq_dyn} at state $\mathbf{x}_i$. The data tuples may be unordered and may originate from multiple trajectories. The dataset index set is denoted by $\mathcal{I}_D = \{1,2,\cdots,M\}$. Throughout this paper, $(\mathbf{x}_t, \mathbf{u}_t)$ denotes a fixed state–input pair, whereas $(\mathbf{x}(t), \mathbf{u}(t))$ represents time-varying variables.

The DKO parameters are obtained by solving the following multi-variable optimization problem over $\mathcal{D}$: \begin{equation}\label{eq_DKO_opti}
    A^*, B^*, C^*, \boldsymbol{\theta}^* = \arg\min_{A,B,C,\boldsymbol{\theta}}\bL_f(A,B,C,\boldsymbol{\theta}), 
\end{equation}
where the loss function is defined as
\begin{equation}
\begin{aligned}
        \bL_f =  \frac{1}{2M}\sum_{i\in\mathcal{I}_D}(\parallel\boldsymbol{g}(\mathbf{x}_i^+,\boldsymbol{\theta}) -A\boldsymbol{g}(\mathbf{x}_i,\boldsymbol{\theta}) -B\mathbf{u}_i \parallel^2 \\+\parallel \mathbf{x}_i^+ - C\boldsymbol{g}(\mathbf{x}_i^+,\boldsymbol{\theta}) \parallel^2). \nonumber
    \end{aligned}
\end{equation} Here, the first term of $\bL_f$ is introduced to approximate \eqref{eq_lift_koopman}, while the second term enforces the reconstruction in \eqref{eq_x_koopman}. At iteration $k$, let $\boldsymbol{\theta}_k$ be the estimation of $\boldsymbol{\theta}^*$. Define the following data matrices constructed from $\mathcal{D}$: \begin{equation}\label{xyudata}
    \begin{aligned}
    \bX &=[\mathbf{x}_1, \mathbf{x}_2,\cdots,\mathbf{x}_M] \in \mathbb{R}^{n \times M},\\
    \mathbf{U} &=[\mathbf{u}_1, \mathbf{u}_2,\cdots,\mathbf{u}_M]\in \mathbb{R}^{m \times M},\\
    \bar{\bX} &= [\mathbf{x}_1^+, \mathbf{x}_2^+,\cdots,\mathbf{x}_M^+]\in \mathbb{R}^{n \times M},\\ 
    \bG_k&= [\boldsymbol{g}(\mathbf{x}_1,\boldsymbol{\theta}_k),\boldsymbol{g}(\mathbf{x}_2,\boldsymbol{\theta}_k),\cdots, \boldsymbol{g}(\mathbf{x}_M, \boldsymbol{\theta}_k)]\in \mathbb{R}^{r \times M},\\
    \bar{\bG}_k&= [\boldsymbol{g}(\mathbf{x}_1^+,\boldsymbol{\theta}_k),\boldsymbol{g}(\mathbf{x}_2^+,\boldsymbol{\theta}_k),\cdots, \boldsymbol{g}(\mathbf{x}_M^+,\boldsymbol{\theta}_k)] \in \mathbb{R}^{r \times M}.
\end{aligned}
\end{equation}
Using \eqref{xyudata}, $\bL_f$ can be rewritten as
\begin{equation}
\begin{aligned}
        \bL_f =  \frac{1}{2M}(\underbrace{\parallel\bar{\bG}_k - [A\ B] \begin{bmatrix}
           \bG_k \\ \mathbf{U} \end{bmatrix}\parallel_F^2}_{\delta(A,B,\boldsymbol{\theta}_k)} + \underbrace{\parallel \bar{\bX} - C\bar{\bG}_k\parallel_F^2}_{\bar{\delta}(C,\boldsymbol{\theta}_k)}). \nonumber
    \end{aligned}
\end{equation}
Under the assumption that the matrices $\bar{\bG}_k \in \mathbb{R}^{r\times M}$ and $[\bG_k', \mathbf{U}']'\in \mathbb{R}^{(r+m)\times M}$ have full row rank (right-invertible), the DKO framework in \cite{hao2025optimalcontrolnonlinearsystems} can be applied to solve \eqref{eq_DKO_opti}:
\begin{equation}\label{eq_gd_thetaf}
    \boldsymbol{\theta}_{k+1} = \boldsymbol{\theta}_k -\alpha_k^f\nabla_{\boldsymbol{\theta}}\bL_f(A_k, B_k, C_k, \boldsymbol{\theta}_k), \quad \boldsymbol{\theta}_0\ \text{given},
\end{equation}
where 
\begin{equation}\label{eq_gd_mats}
    \begin{aligned}
        [A_k\ B_k] &= \arg\min_{[A\ B]} \delta(A,B,\boldsymbol{\theta}_k) = \bar{\bG}_k \begin{bmatrix} \bG_k \\ \mathbf{U} \end{bmatrix}^\dagger, \\
        C_k &= \arg\min_{C}\bar{\delta}(C,\boldsymbol{\theta}_k)=  \bar{\bX}\bar{\bG}_k^{\dagger},
    \end{aligned}
\end{equation} are constant matrices determined by $\boldsymbol{\theta}_k$. The convergence properties of \eqref{eq_gd_thetaf}-\eqref{eq_gd_mats} are analyzed in \cite{hao2025optimalcontrolnonlinearsystems}.
% \begin{algorithm}
% \SetKwInOut{Input}{Input}
% \Input{Task, Episodes $E$, Episode steps $M$,
% Initial dataset $\mathcal{D}$, iteration $k=0$, Initial control sequence $(\mathbf{u}_0, \mathbf{u}_1, ... \mathbf{u}_{T-1})$\;}
% \For{$i \leftarrow 1$ \KwTo $E$}{
%   Update $\boldsymbol{f}_k$ in \eqref{eq_approx_dko_update} following \eqref{eq_gd_thetaf} using $\mathcal{D}$ \;\For{$j \leftarrow 1$ \KwTo $M$}{$\mathbf{u}_j\leftarrow \textrm{MPPI}(\boldsymbol{f}_k, Task)$ \; Execute $\mathbf{u}_j$ and observe new state $\mathbf{x}_j^+$\; $\mathcal{D}\leftarrow \mathcal{D}\cup\{(\mathbf{x}_j, \mathbf{u}_j, \mathbf{x}_j^+)\}$\;} $k+=1$\;
% }
% % \Return $\{w_1, w_2, \dots w_K \}$
% \caption{Deep Koopman Learning} 
% \label{alg: dko}
% \end{algorithm}

\subsection{MPPI control}
The classic MPPI controller solves the stochastic optimal control problem in \eqref{eq_dk_mppi_optim} by enforcing the true system dynamics in \eqref{eq_dyn} instead of the Koopman dynamics constraint. At each time step $t$, MPPI operates in a receding-horizon manner, performing Monte Carlo rollouts to generate $N$ sampled trajectories, each propagated forward using the system dynamics over a finite prediction horizon $T$. Let $n=1,2,\cdots,N$ denote the index of sampled trajectories. The cumulative cost over the finite prediction horizon $T$ (the cost-to-go) of the $n$-th sampled trajectory is defined as
\begin{equation}
    S_n = \phi(\mathbf{x}_{t+T}) + \sum_{s=t}^{t+T-1} \hat{c}(\mathbf{x}_s, \mathbf{u}_s, \delta\mathbf{u}_s^n), 
\end{equation}
where $\delta\mathbf{u}_s^n$ denotes zero-mean Gaussian noise corresponding to $\mathbf{u}_s$ at the $n$-th trajectory and $\hat{c}$ denotes the instantaneous running cost. The running cost $\hat{c}$ consists of a state-dependent cost $c$ and a quadratic control penalty, and is given by
\begin{equation}
\begin{aligned}
    \hat{c} = \frac{1}{2}\mathbf{u}_s'R\mathbf{u}_s + \gamma(\delta\mathbf{u}_s^n)'R\delta\mathbf{u}_s^n + \mathbf{u}_s'R\delta\mathbf{u}_s^n + c(\mathbf{x}_s),
\end{aligned}
\end{equation}
where $\gamma = \frac{\nu-1}{2\nu}\in\mathbb{R}_+$ with $\nu \geq 1$ regulating the exploration–exploitation trade-off.

Following the classic MPPI update rule \cite{williams2017model}, the nominal control sequence $\{\mathbf{u}_s\}_
{s=t}^{t+T-1}$ is updated using a cost-weighted average of the sampled perturbations:
\begin{equation}\label{eq_standard_mppi}
    \mathbf{u}_s\leftarrow \mathbf{u}_s + \frac{
    \sum_{n=1}^N \exp\Big(\frac{-1}{\lambda} \Big[S_n - S_{\mathrm{min}}\Big]\Big) \delta\mathbf{u}_s^n
    }{
    \sum_{n=1}^N \exp\Big(\frac{-1}{\lambda} \Big[S_n - S_{\mathrm{min}}\Big]\Big)
    },
\end{equation}
where $S_{\mathrm{min}} = \min_n S_n$ is the minimum cost among all rollouts, introduced to improve numerical stability, and $\lambda$ is the inverse temperature parameter governing the selectiveness of the weighting scheme.

After applying the update rule in \eqref{eq_standard_mppi} at each iteration, the resulting control sequence is optionally smoothed using a Savitzky–Golay filter \cite{savitzky1964smoothing}. In a receding-horizon manner, only the first control input $\mathbf{u}(t)$ is applied to the system, while the remaining sequence is used as a warm start for the subsequent optimization step.

\section{Methodology}\label{MaRe}
This section presents an iterative procedure for solving \eqref{eq_dk_mppi_optim}. At any time $t$, the learned DKO dynamics are expressed as \begin{equation}\label{eq_approx_dko_update}
\begin{aligned}
    \mathbf{\hat x}(t+1) = C^* \Big(A^* \boldsymbol{g}(\mathbf{\hat x}(t),\boldsymbol{\theta}^*) + B^* \mathbf{u}(t)\Big), \quad \mathbf{\hat x}(t) = \mathbf{x}(t), \nonumber
\end{aligned}
\end{equation} 
where $A^*$, $B^*$, $C^*$, $\boldsymbol{\theta}^*$ are obtained from the DKO updating rule in \eqref{eq_gd_thetaf}-\eqref{eq_gd_mats} using the training dataset $\mathcal{D}$. In this work, $\mathcal{D}$ is collected by applying uniformly sampled control inputs to the true dynamical system. In practice, the dataset may also be gathered using an MPPI controller initialized with a preliminary DKO model or from demonstrations provided by a human operator in real-world experiments.

Based on the learned linear DKO dynamics, we develop an MPPI controller built upon the DKO model, referred to as MPPI with deep Koopman operator dynamics (MPPI-DK), which is summarized in Algorithm \ref{alg: mppi-dk}.
\begin{algorithm}
\SetKwInOut{Input}{Given}
\Input{Sampling horizon: $T$\;Number of sampling trajectories: $N$\;
      Cost functions/parameters:  $c, \phi, \Sigma, \lambda$ \; Learned dynamics matrices $A^*$, $B^*$, $C^*$\;
Current state $\mathbf{x}_t$ and its lifted vector $\boldsymbol{g}(\mathbf{x}_t,\boldsymbol{\theta}^*)$ \; Control sequence: $(\mathbf{u}_0, \mathbf{u}_1, ... \mathbf{u}_{T-1})$\;}        
\While{task not completed}{
\For{$n \leftarrow 0$ \KwTo $N-1$}{
  $\mathbf{x} \leftarrow \mathbf{x}_t$,  $\mathbf{g} \leftarrow\boldsymbol{g}(\mathbf{x}_t,\boldsymbol{\theta}^*)$,
  $S_n \leftarrow 0$\;
  Sample $[\delta\mathbf{u}_0^n \dots \delta\mathbf{u}_{T-1}^n], ~\delta\mathbf{u}_s^n \sim \mathcal{N}(0, \Sigma)$\;
  \For{$s \leftarrow 1$ \KwTo $T$}{
    $\mathbf{x} \leftarrow C^*(A^* \mathbf{g} + B^*(\mathbf{u}_{s-1}+\delta\mathbf{u}_{s-1}^n))$ \;
    $\mathbf{g}\leftarrow A^*\mathbf{g} + B^*(\mathbf{u}_{s-1}+\delta\mathbf{u}_{s-1}^n)$\;
  	$S_n \pluseq  c(\mathbf{x}) + \gamma \mathbf{u}_{s-1}^T \Sigma^{-1} \delta\mathbf{u}_{s-1}^n$\;
  }
  $S_n \pluseq \phi(\mathbf{x})$\;
}
$S_{\mathrm{min}} \leftarrow \min_n S_n$\;
\For{$s \leftarrow 0$ \KwTo $T-1$}{
$\mathbf{u}_s += \frac{\sum_{n=0}^{N-1} \exp(\frac{-1}{\lambda}(S_n- S_{\mathrm{min}}))\delta\mathbf{u}_s^n}{\sum_{n=0}^{N-1}\exp(\frac{-1}{\lambda}(S_n - S_{\mathrm{min}}))}$\;
}

$\text{SendToActuators}(\mathbf{u}_0)$\;

\For{$s \leftarrow 1$ \KwTo $T-1$}{ 
	$\mathbf{u}_{s-1} \leftarrow \mathbf{u}_s$\;  
}
$\mathbf{u}_{T-1} \leftarrow \text{Initialize}(\mathbf{u}_{T-1})$\;
}
\caption{MPPI with Learned Deep Koopman Operator Dynamics (MPPI-DK)}
\label{alg: mppi-dk}
\end{algorithm}
\begin{remark}
   During the trajectory sampling process in the proposed MPPI-DK algorithm, once the state $\mathbf{x}$ is updated, the corresponding lifted state $\boldsymbol{g}(\mathbf{x},\boldsymbol{\theta}^*)$ is propagated using the learned linear DKO dynamics matrices $A^*$ and $B^*$, rather than being recomputed through the DNN $\boldsymbol{g}(\mathbf{x},\boldsymbol{\theta}^*)$. This substitution reduces the computational burden and improves sampling efficiency, particularly when $\boldsymbol{g}$ is complex.
\end{remark}

\section{Numerical Simulations}\label{numerical_sim}
This section first evaluates the proposed method on a simulated inverted pendulum task to examine the effects of key parameters on control performance. It is then validated on a surface-vehicle navigation task using classic MPPI with the true system dynamics as a benchmark, enabling assessment of the computational gains from linear DKO propagation and the performance loss due to model approximation. All DNNs are trained using the Adam optimizer \cite{kingma2014adam}.

\subsection{Pendulum Swing-Up and Stabilization}
In this task, the controller needs to swing up the pendulum from an arbitrary initial state and stabilize it at the upright equilibrium. We examine how the training data and the architecture of the DNN lifting function $\boldsymbol{g}$ in \eqref{eq_approx_dko}, including the number of neurons and lifted state dimension, affect MPPI-DK performance. The pendulum dynamics are given by
\begin{equation}\label{eq_pen_dyn}
    \begin{cases}
    \psi(t+1) = \psi(t) + \dot{\psi}(t+1) \Delta t,\\
      \dot{\psi}(t+1) = \dot{\psi}(t) + (\frac{-3g\sin(\psi(t) + \pi)}{2l} + \frac{3u(t)}{ml^2})\Delta t, 
\end{cases}
\end{equation} where $g=10\mathrm{m/s^2}$, $m=1\mathrm{kg}$, and $l=1\mathrm{m}$ represent the gravitational acceleration, the mass of the pendulum, and the length of the pendulum, respectively, and $\Delta t = 0.05\mathrm{s}$ denotes the discretization step. Here, the system state is defined as $\boldsymbol{x}(t) = [\psi(t), \dot{\psi}(t)]'$ with the lower bound and upper bound of $[-\pi,-8]'$ and $[\pi,8]'$, respectively, and $u(t)$ denotes the continuous scalar torque input, constrained between $-2$ and $2$. The state-dependent cost function is defined as $$c(\boldsymbol{x}(t)) = \psi(t)^2 + 0.1\dot{\psi}(t)^2.$$ For each episode, the initial state $\boldsymbol{x}(0)$ is uniformly sampled from the interval between $[-\pi,-1]'$ and $[\pi,1]'$, and the episode is terminated when $t>200$.  

\textbf{Setup.} (1) \textit{Training data}: To evaluate the effect of training data on the performance of the proposed MPPI-DK controller, we construct two training datasets. The first dataset is generated by applying control inputs uniformly sampled from the interval $[-2, 2]$. The second dataset augments the first by incorporating expert demonstrations obtained from an MPC controller capable of completing the pendulum swing-up task, with an expert-to-non-expert data ratio of $1:1$. (2) \textit{Lifting functions}: We further investigate the control performance of MPPI-DK under different architectures of the lifting function $\boldsymbol{g}$. The evaluated network structures, including the hidden layer sizes and the lifted state dimension (i.e., the output dimension of the DNN), are summarized in Table \ref{DNN-table}.
\begin{table}[ht]
\centering
\begin{tabular}{l|c|c|c|c}
\hline
 layer $1$ size & layer $2$ size & hidden layers & output layer& lift dim \\
 \hline
 $16$ & $16$ & $ReLU$&$Tanh$&$4$\\
 % \hline
 $32$ & $32$ & $ReLU$&$Tanh$&$4$\\
 % \hline
 $64$ & $64$ & $ReLU$&$Tanh$&$4$\\
 % \hline 
 $64$ & $64$ & $ReLU$&$Tanh$&$6$\\
 % \hline 
 $64$ & $64$ & $ReLU$&$Tanh$&$8$\\
 \hline
\end{tabular}
\caption{\label{DNN-table} The structure of $\boldsymbol{g}$.}
\end{table}

To ensure a fair comparison, we evaluate the state-dependent costs of trajectories initialized from a challenging state $\mathbf{x}_0=[\pi, 0.1]'$ (downward) and converging to the target state $[0,0]'$. The results are compared across different training datasets and lifting function architectures, as well as against classic MPPI using \eqref{eq_pen_dyn} as a benchmark. Both methods use identical parameters: $T = 20$, $N = 2000$, $\Sigma = 1.0$, $\lambda = 0.1$. Although this specific initial condition is chosen for evaluation, the proposed method is applicable to arbitrary initial states. The experiment is repeated over $5$ independent trials to mitigate the effects of randomness in DNN training.

\textbf{Results analysis.} 
\begin{figure}[ht]
    \centering
\includegraphics[width=0.82\linewidth]{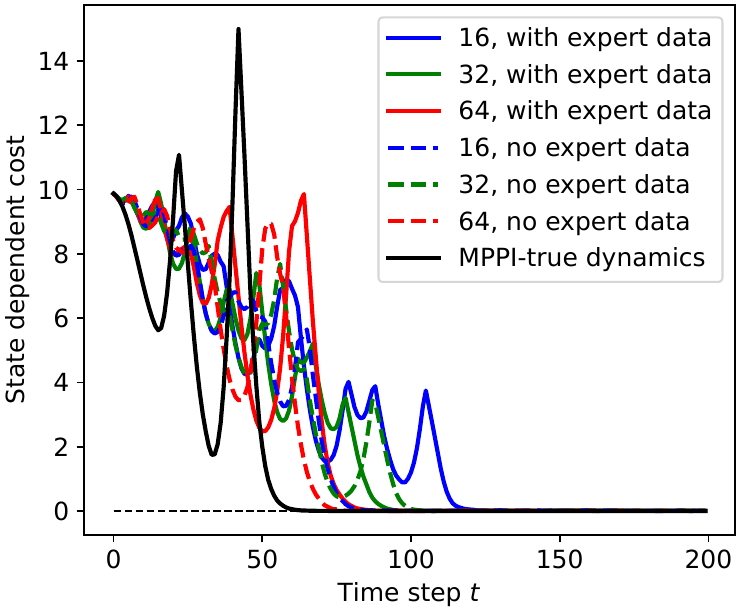}
    \caption{Average trajectory cost over $5$ independent trials for different lifting function architectures and training datasets.}
    \label{fig:cost_num}
\end{figure}
\begin{figure}[ht]
    \centering
\includegraphics[width=0.83\linewidth]{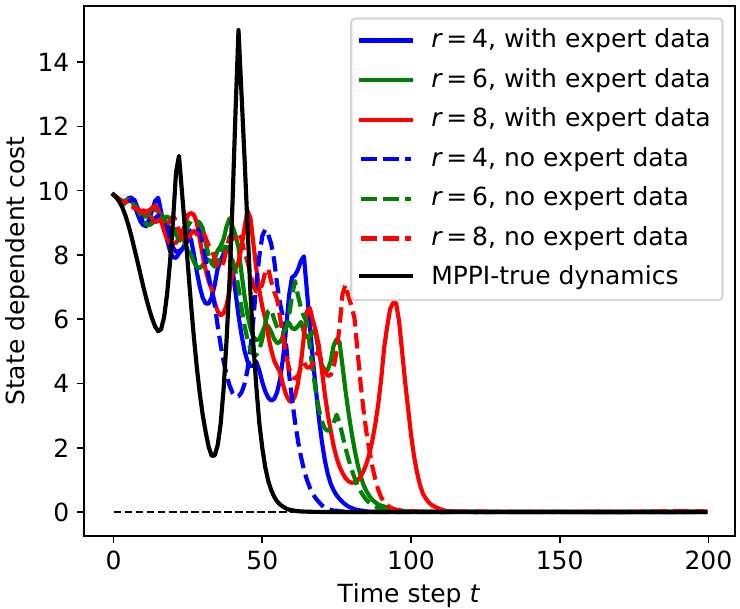}
    \caption{Average trajectory cost over $5$ independent trials for different lifting dimensions and training datasets.}
    \label{fig:cost_dim}
\end{figure}
% \begin{figure}[ht]
%     \centering
% \includegraphics[width=0.83\linewidth]{figs/err_long_num.pdf}
%     \caption{Long-term propagation errors. Solid and dashed lines indicate the average over $5$ independent trials, and the shaded region denotes the standard deviation.}
%     \label{fig:prop_err}
% \end{figure}
As shown in Figs.~\ref{fig:cost_num}–\ref{fig:cost_dim}, increasing the number of neurons results in faster convergence of the MPPI-DK controller to the goal state, with control inputs becoming more aggressive and trajectories more closely matching those generated by classic MPPI using the true system dynamics. In contrast, increasing the lifting dimension or augmenting the training set with expert demonstrations does not consistently improve control performance in the pendulum swing-up task.

\subsection{Surface Vehicle Navigation}
For this task, the controller is designed to steer the surface vehicle toward a desired goal position. The acceleration dynamics of the simulated surface vehicle are given by \begin{equation}\label{eq_boat_dyn}
\mathbf{\dot s} = M \boldsymbol{\phi}(\mathbf{s}, \mathbf{u})
\end{equation}
where $\mathbf{s} = \begin{bmatrix}v_x, v_y, \dot{\psi}\end{bmatrix}'\in\mathbb{R}^3$ denotes the body-frame linear velocities and angular velocity, respectively, $\mathbf{u} = \begin{bmatrix}u_{\mathrm{left}},  u_{\mathrm{right}}\end{bmatrix}'\in\mathbb{R}^2$ represents the thrusts of the left and right motors constrained between $[-1,-1]'$ and $[1,1]'$, $M\in\mathbb{R}^{3\times 22}$ is a constant matrix and $\boldsymbol{\phi}:\mathbb{R}^3\times\mathbb{R}^2\rightarrow\mathbb{R}^{22}$ is a nonlinear mapping defined as $\boldsymbol{\phi}(\mathbf{s}, \mathbf{u}) =
[v_x, v_y, \dot{\psi}, v_x v_y, v_x \dot{\psi}, v_y \dot{\psi},
v_x^2, v_y^2, \dot{\psi}^2,
v_x |v_x|, v_x |v_y|, v_x |\dot{\psi}|,\\
v_y |v_x|, v_y |v_y|, v_y |\dot{\psi}|,
\dot{\psi} |v_x|, \dot{\psi} |v_y|, \dot{\psi} |\dot{\psi}|,
u_{\mathrm{left}}, f_2, u_{\mathrm{right}}, f_4]'$. Here, for all $t$, $f_2 = f_4 =0$, since the rudder angle is assumed to be fixed. The dynamics in \eqref{eq_boat_dyn} is simulated using Euler discretization $ \mathbf{s}(t+1) = \mathbf{s}(t) + \mathbf{\dot s}(t)\Delta t,$ where $\Delta t >0$ is the discretization step. Using the dynamics in \eqref{eq_boat_dyn}, the corresponding kinematic model is given by \begin{equation}\label{eq_boat_kin}
\mathbf{\dot{p}}(t) = \begin{bmatrix}
    \cos(\psi(t))v_x(t+1) - \sin(\psi(t))v_y(t+1) \\
    \sin(\psi(t))v_x(t+1) + \cos(\psi(t))v_y(t+1) \\ \dot{\psi}(t+1)
\end{bmatrix},
\end{equation} where $\mathbf{p}(t) = \begin{bmatrix}x(t), y(t), \psi(t)\end{bmatrix}'\in\mathbb{R}^3$ denotes the global position and yaw angle, and it is simulated using the Euler discretization $\mathbf{p}(t+1) = \mathbf{p}(t) + \mathbf{\dot{p}}(t)\Delta t$. 

\textbf{Setup.} The complete state is $\mathbf{x}(t)=[\mathbf{p}(t)',\mathbf{s}(t)']'$. The vehicle is driven from $\mathbf{x}_0=[20,10,\pi/3,0,0,0]'$ to $\mathbf{x}_{\mathrm{goal}}=[0,0,\pi/2,0,0,0]'$ and the cost function is defined as
$c(\mathbf{x}(t))=\parallel\mathbf{x}(t)-\mathbf{x}_{\mathrm{goal}}\parallel^2.$
Because the kinematics follow directly from \eqref{eq_boat_dyn}, the DKO model learns only the mapping $(\mathbf{s}(t),\mathbf{u}(t))\mapsto\mathbf{s}(t+1)$. The lifting function $\boldsymbol{g}$ follows Table~\ref{DNN-table}, with $256$ neurons per hidden layer and lifting dimension $8$. Training tuples $(\mathbf{s}_t,\mathbf{u}_t,\mathbf{s}_t^+)$ are generated using uniformly sampled inputs $\mathbf{u}_t$ from $[-1,-1]'$ to $[1,1]'$.

To evaluate computational efficiency, MPPI-DK is compared with classic MPPI using the true dynamics and with MPC using the same learned DKO model, thereby isolating the effects of model approximation. MPPI-DK and classic MPPI use identical costs and parameters: $T=20$, $N=600$, $\Sigma=\operatorname{diag}(0.8,0.8)$, and $\lambda=20$. The MPC horizon is also $T=20$, and its optimization problem is solved using IPOPT through CasADi. MPPI-DK is additionally implemented on a GPU for parallel trajectory sampling. All experiments are repeated over $4$ independent DNN training trials.

\textbf{Results analysis.} 
\begin{figure}[ht]
    \centering
\includegraphics[width=0.83\linewidth]{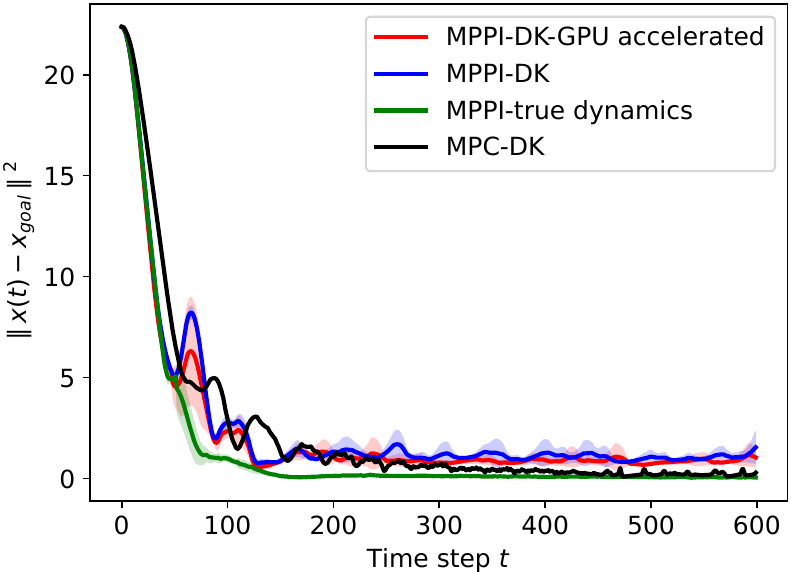}
    \caption{Tracking error over time. The solid line represents the mean over $4$ independent trials, and the shaded region indicates one standard deviation.}
    \label{fig:errors}
\end{figure}
% \begin{figure}[ht]
%     \centering
% \includegraphics[width=0.83\linewidth]{figs/compute_cost.pdf}
%     \caption{Per-step computation time (log scale) over time. The solid line denotes the mean over $4$ independent trials, and the shaded region indicates one standard deviation.}
%     \label{fig:com}
% \end{figure}
% \begin{table}[t]
% \centering
% \setlength{\tabcolsep}{3pt}
% \begin{tabular}{lcccc}
% \toprule
% Method & Proposed (GPU) & Proposed (CPU) & MPPI (True) & MPC-DK  \\
% \midrule
% Time (ms) & $17.9\pm 1.0$ & $332.0\pm 7.7$ & $2041.7\pm 219.3$ & $244.3$ \\
% \bottomrule
% \end{tabular}
% \caption{Per-step computation time (milliseconds).}
% \label{table:com}
% \end{table}
\begin{table}[t]
\centering
\setlength{\tabcolsep}{4pt}
\begin{tabular}{lccc}
\toprule
Method & Dynamics model & Device & Time (ms) $\downarrow$ \\
\midrule
\textbf{Proposed}   & DKO & GPU & $17.9 \pm 1.0$ \\
\textbf{Proposed}   & DKO & CPU & $332.0 \pm 7.7$ \\
Classic MPPI & True dynamics & GPU & $34.4 \pm 27.7$ \\
Classic MPPI & True dynamics & CPU & $2041.7 \pm 219.3$ \\
MPC-DK    &  DKO & CPU & $87.1$ \\
\bottomrule
\end{tabular}
\caption{Average per-step computation time for the surface-vehicle task with $T=20$. GPU results (NVIDIA RTX 3080) demonstrate the benefit of parallel sampling, while CPU results (Apple M2) provide a hardware-matched comparison.}
\label{table:com}
\end{table}
As shown in Fig.~\ref{fig:errors}, the proposed approach achieves tracking performance close to classic MPPI with access to the true system dynamics. The computational performance is summarized in Table~\ref{table:com}. When executed on a CPU, MPPI-DK requires less computation time per control step than the classic MPPI, although it remains more computationally intensive than MPC based on the same learned DKO model. With GPU-enabled parallel sampling, MPPI-DK improves computational efficiency by approximately $50$\% relative to classic MPPI using the true dynamics and is substantially more efficient than MPC implemented with the identical DKO model. During the simulations, the MPC solver occasionally encountered infeasible optimization problems due to the imposed dynamics constraints, resulting in a marked increase in computation time.

\section{Experimental Results}\label{exp}
We evaluate the proposed approach on a reference-tracking task using the quadruped robot Unitree Go1 platform \cite{unitree_go1}. The quadruped robot is modeled with state \[\mathbf{x} = [x,y,\psi,v_x,v_y,\dot{\psi}, \mathbf{h}']'\in\mathbb{R}^9,\] where $x$ and $y$ denote the robot global position, $\psi$ is the yaw angle,  $v_x$, $v_y$ are the body-frame linear velocities obtained from the world-frame velocities by $$\begin{bmatrix} v_x \\ v_y \end{bmatrix}
= \begin{bmatrix}
\cos\psi & \sin\psi\\
-\sin\psi & \cos\psi
\end{bmatrix}\begin{bmatrix} \dot{x} \\ \dot{y} \end{bmatrix}.$$ The centroidal momentum state \cite{orin2013centroidal, dai2014whole} is defined as $\mathbf{h} = [l_x, l_y, w_m]'\in\mathbb{R}^3$, where $l_x=m\dot{x}$, $l_y=m\dot{y}$ denote the linear momenta about the center of mass (CoM), and $w_m = I_{zz}\dot{\psi}$ denotes the angular centroidal momentum about the CoM. Here, $m=12\mathrm{kg}$ is the robot mass, and $I_{zz} = 0.9 \mathrm{kg\cdot m^2}$ is the yaw moment of inertia about the CoM. The control input \[\mathbf{u}=[a_{\mathrm{fwd}}, a_{\mathrm{lat}}, a_{\mathrm{z}}]'\in\mathbb{R}^3\] denotes the commanded forward and lateral accelerations in the body frame, and the desired yaw angular acceleration, respectively. The control input is constrained between $[-1,-1,-1]'$ and $[1,1,1]'$.
The centroidal momentum dynamics are given by
\[
\mathbf{\dot h} =
\begin{bmatrix}
m\cos\psi & -m\sin\psi & 0\\
m\sin\psi & \;\;m\cos\psi & 0\\
0 & 0 & I_{zz}
\end{bmatrix}
\mathbf{u}.
\]
Let $\mathbf{p}_{\mathrm{com}} = [x, y, z]$ denote the CoM position in the global frame with constant height $z$. The CoM kinematics follow $\mathbf{\dot p}_{\mathrm{com}}=[l_x/m, l_y/m, 0]'$ and $\dot\psi=w_m/I_{zz}$.

\textit{Task-Oriented State Definition.} The control objective is to drive the quadruped to the desired goal pose $[x_{\textrm{goal}}, y_{\textrm{goal}},\psi_{\textrm{goal}}]' = [1.5, 0, 0]'$. Since the task primarily depends on the base motion, we adopt an $8$-dimensional task-oriented state defined as \begin{equation}\label{eq_task_s}
    \mathbf{s} = [\Delta x, \Delta y, \Delta\psi, \cos\psi, \sin\psi,
v_x,v_y, \dot{\psi}]'\in \mathbb{R}^{8},
\end{equation} where $\Delta x=x - x_{\textrm{goal}}$, $\Delta y= y - y_{\textrm{goal}}$, and $\Delta\psi = \psi-\psi_{\textrm{goal}}$.  The state-dependent cost is defined as \[c(\mathbf{s}) = 6\Delta x^2 + 20\Delta y^2 + 4.5\Delta \psi^2.\] 

\textbf{Setup.}
To learn the DKO dynamics for the mapping between $(\mathbf{s}(t), \mathbf{u}(t))$ and $\mathbf{s}(t+1)$, we generate training tuples $(\mathbf{s}_t, \mathbf{u}_t,\mathbf{s}_t^+)$ by applying uniformly sampled control inputs to the quadruped system. The DKO lifting network is constructed following the architecture specified in Table \ref{DNN-table}, with $256$ neurons per layer and a lifting dimension of $10$. We compare MPPI-DK with classic MPPI using the true robot dynamics. Both controllers employ GPU-parallel trajectory sampling with identical parameters: $T=40$, $N=900$, $\Sigma=\operatorname{diag}(0.18,0.18,0.18)$, and $\lambda=0.5\operatorname{std}(S)$, where $\operatorname{std}(S)$ is the standard deviation of the sampled cumulative costs at each control step. This adaptive temperature improves numerical stability and balances exploration and exploitation. Table~\ref{Comparison-table} reports the per-step computation time, average trajectory-tracking error, and final tracking error over $10$ initial states.

\begin{figure}[t]
    \centering
    \includegraphics[width=\linewidth]{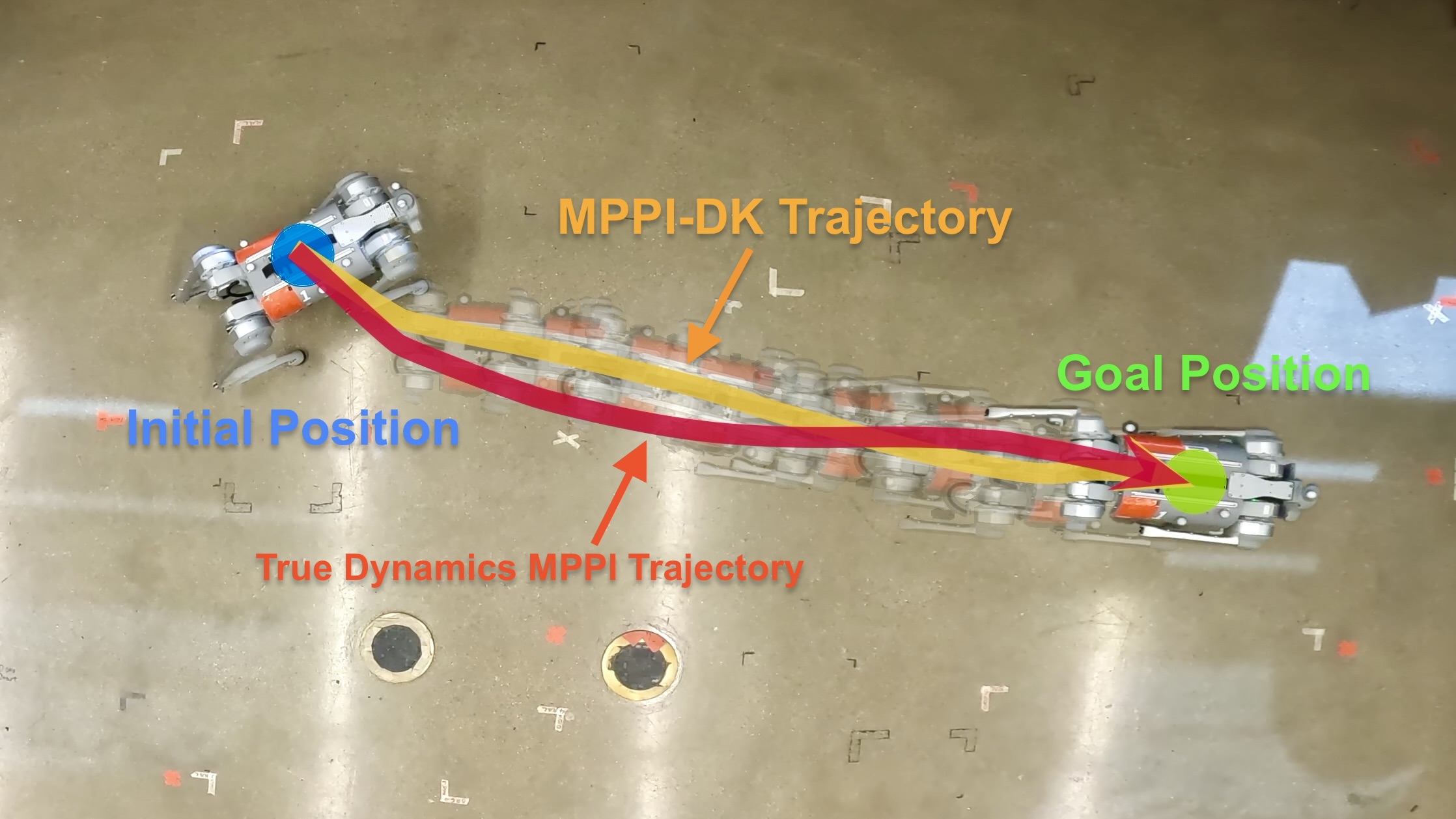}
    \caption{Representative Unitree Go1 trajectories generated by MPPI-DK and classic MPPI using the true dynamics, from $[x, y, \psi] = [-0.5 \text{m}, 0.4 \text{m}, 0.3 \text{rad}]'$ to $[1.5 \text{m}, 0 \text{m}, 0 \text{rad}]'$.}
    \label{fig:header}
\end{figure}

\textbf{Results analysis.}
\begin{table}[t]
\centering
\setlength{\tabcolsep}{3pt}
\begin{tabular}{lcccc}
\toprule
Method & Time (ms)$\downarrow$ & Track Err$\downarrow$ & Final Err$\downarrow$ & Smooth$\downarrow$  \\
\midrule
\textbf{Proposed}  & $8.8\pm0.1$ & $0.658\pm0.030$ & $0.004\pm0.001$ & $20.6\pm 9.9$ \\
MPPI-True & $11.7\pm0.1$ & $0.534\pm0.002$ & $0.007\pm0.001$& $26.4\pm 13.3$ \\
\bottomrule
\end{tabular}
% \caption{Performance comparison over $10$ initial states (GPU implementation). Metrics include per-step computation time, average and final tracking errors, and control smoothness $V = \frac{1}{T}\sum_{t=0}^{T-1} \parallel \mathbf{u}(t+1) -\mathbf{u}(t)\parallel^2$}
\caption{GPU-based (NVIDIA RTX 3080) performance over $10$ initial states, with control smoothness $V = \frac{1}{T-1}\sum_{t=0}^{T-2} \parallel \mathbf{u}(t+1) -\mathbf{u}(t)\parallel^2$.}
\label{Comparison-table}
\end{table}
% \begin{figure}[ht]
%     \centering
% \includegraphics[width=0.83\linewidth]{figs/track_err_dog.pdf}
%     \caption{Tracking error over time, averaged across $10$ different initial states. The solid line denotes the mean value, and the shaded region indicates one standard deviation.}
%     \label{fig:dog_tracing_err}
% \end{figure}
As shown in Table~\ref{Comparison-table}, both controllers complete the task from all $10$ initial states. MPPI-DK requires less computation time per step, achieves slightly lower final tracking error, and produces smoother control inputs. A representative trajectory is shown in Fig.~\ref{fig:header}.

\section{Concluding Remarks}\label{conclusion}
We introduced MPPI-DK, a DKO-dynamics-based MPPI framework that improves trajectory-sampling efficiency by replacing nonlinear rollout propagation with learned linear Koopman dynamics. The method was evaluated on inverted pendulum and surface-vehicle tasks, achieving control performance comparable to MPPI using the true dynamics at lower computational cost, and was further validated through quadruped reference-tracking experiments. The hardware results similarly demonstrate accurate tracking and confirm the practical potential of structured Koopman models for efficient sampling-based control of nonlinear robotic systems. Future work will investigate online adaptation of the Koopman model, uncertainty-aware trajectory sampling, and extensions to more complex robotic systems and safety-critical tasks.

\bibliographystyle{unsrt}
\bibliography{refs_hao.bib}

\end{document}